\documentclass[journal]{IEEEtran}

\ifCLASSINFOpdf
\else
\fi
\usepackage{natbib}

\newcommand{\bd}{\begin{displaymath}}
\newcommand{\ed}{\end{displaymath}}
\newcommand{\be}{\begin{equation}}
\newcommand{\ee}{\end{equation}}
\newcommand{\bea}{\begin{eqnarray}}
\newcommand{\eea}{\end{eqnarray}}
\newcommand{\ba}{\begin{array}}
\newcommand{\ea}{\end{array}}
\newcommand{\bc}{\begin{center}}
\newcommand{\ec}{\end{center}}

\newcommand{\nb}{\nonumber}

\newcommand{\re}[1]{(\ref{#1})}

\usepackage[colorlinks,bookmarksopen,bookmarksnumbered,citecolor=red,urlcolor=red]{hyperref}
\usepackage{graphicx}

\begin{document}
%
\title{Intention assimilation control for accurate tracking with variable impedance in teleoperation}
%
%
%

\author{Atsushi Takagi*, Yanan Li*, Hiroaki Gomi and Etienne Burdet%
\thanks{*These authors contributed equally to the work.}}

%
%

\markboth{}%
{Shell \MakeLowercase{\textit{et al.}}: Bare Demo of IEEEtran.cls for IEEE Journals}
%



\maketitle
\begin{abstract}
Robot systems for teleoperation commonly use a spring-like force pulling the follower robot towards the leader's position to track their movements. With this control strategy, the tracking accuracy deteriorates when the follower' stiffness is low, but high stiffness poses a danger to objects or people in the follower robot's environment. To address this trade-off between tracking accuracy and safety, we propose an alternative intention assimilation control (IAC) strategy where the robot's tracking accuracy can be ensured without high stiffness. Different from traditional approaches, which transmit the leader's current position to the follower, this new controller estimates the leader's target position and transmits it to the follower. With this strategy, the follower impedance can be changed on-the-fly to continuously reflect the user's desired impedance or modulated automatically to fulfill the task requirements. Our controller was validated on two 7 degree-of-freedom manipulators, yielding high tracking accuracy with varying impedance. Four experiments were conducted to compare {teleoperation} with IAC to tele-impedance control (TIC) during free tracking, interaction with a balloon, during peg insertion, and table polishing with force feedback. The results show that IAC increases tracking accuracy, improves task completion rate and reduces completion time. IAC enables the robot to accurately replicate the user's movement while giving them freedom to modulate the impedance according to their intention, providing an unprecedented level of control of the follower's position and its impedance during unilateral and bilateral teleoperation.
\end{abstract}


%
\IEEEpeerreviewmaketitle

\section{Introduction}
Teleoperation has been extensively studied in the literature \citep{PassenbergC10M, HircheS12PI, yokokohjiBilateralControlMasterslave1994} and used in various application domains including surgery \citep{Tobergte09, Xia2011}, physical rehabilitation \citep{Atashzar2012} and space exploration \citep{Yoon2004, panzirschExploringPlanetGeology2022}. A typical teleoperation system consists of a robotic interface operated by a human user leader and a follower robot controlled remotely and interacting with its environment. {Teleoperation can be unilateral, where the leader receives no force feedback from the follower robot, or bilateral where the force experienced by the follower is transferred back to the leader. The communication between the leader and follower robots is affected by delays in the transmission signals which can range from a few milliseconds for local setups (e.g. surgery where operator and robot are in the same room) to minutes in space exploration applications. 

Communication delays can make bilateral teleoperation unstable due to energy injection from wave reflections. Various techniques have been developed to address this issue, such as using wave-variables \citep{andersonBilateralControlTeleoperators1988, niemeyerStableAdaptiveTeleoperation1991}, four channel architecture \citep{lawrenceStabilityTransparencyBilateral1993}, model-based method \citep{fundaTeleprogrammingDelayinvariantRemote1992}, or predictive control \citep{panNewPredictiveApproach2006}. Unilateral teleoperation, where commands go solely from the leader to the follower robot, does not provide the human user with haptic percepts of the environment, but it is robust to communication delays. However, the above-mentioned bilateral and unilateral teleoperation techniques fail to match the follower's impedance to the leader's, which can be problematic for interaction tasks like polishing and insertion where humans continuously adapt their arm's impedance in accordance with the task \citep{ajoudaniTeleimpedanceTeleoperationImpedance2012, takagiAnalogousAdaptationsSpeed2020}. 

This problem was addressed by a state-of-the-art teleoperation scheme dubbed tele-impedance control (TIC) \citep{ajoudaniTeleimpedanceTeleoperationImpedance2012, Laghi2020}, which estimates the impedance of the arm of the operator based on the estimated arm's muscle activity. With TIC} the follower's position is pulled towards a reference trajectory by a controller whose impedance is set equal to that of the human arm and thus reflected onto the remote environment. A shortcoming of TIC is that the follower can track the leader's trajectory accurately only when the leader's movement is slow, or when the leader's impedance is high, which may make the follower unsafe in its operation. Thus, the operator has to compromise either tracking accuracy or safety, which is undesirable in settings where the follower interacts or moves in a space shared with humans.

Our goal is to develop a teleoperation strategy that allows the human operator to modulate the follower's impedance broadly from low to high stiffness without compromising the follower's accuracy in tracking the leader's position. Our idea is inspired by recent neuroscience studies which show that humans use haptic information to predict a partner's movements during physical interactions \citep{Ganesh2014, Takagi2017, Takagi2018a}. The proposed controller infers the human operator's movement intention from the force they apply on the leader robot as in \citep{gomiEquilibriumPointControlHypothesis1996}, and uses it to plan the follower robot's movement. Related studies have used haptic information from the partner to improve human-robot interaction \citep{Frizera2010, Li2014, Li2019, Takagi-Li2020}, but to our knowledge this has never been applied to teleoperation.

The \textit{intention assimilation control} (IAC) method introduced in this paper is implemented through the following steps:
\begin{itemize}
\item The follower robot estimates the human operator's target trajectory from the force applied onto the leader device \citep{Takagi2017, Takagi-Li2020}.
\item The follower robot's impedance is controlled by the operator by estimating their arm's impedance magnitude using their power grasp force \citep{takagiEndpointStiffnessMagnitude2020}.
\item The estimated leader's target trajectory and impedance are used by the follower robot to proactively follow the leader. During bilateral teleoperation, the follower's interaction torque is fed back to the leader.
\end{itemize}

While both TIC and IAC use impedance identified at the operator to control the follower robot, there is a key difference in the information sent from the leader to the follower robot. In TIC, the leader's trajectory and impedance are sent to the follower. In IAC, the \emph{leader's target trajectory} and impedance are sent. Therefore, IAC allows the follower robot to use a low or variable impedance in scenarios of uncertain environments to ensure safe interactions without compromising the follower's tracking accuracy, while TIC can only achieve a trade-off between interaction safety and tracking performance. 

We compared the performance of these two controllers by carrying out free motion and interaction tasks using two 7 DoF manipulators { with a constant communication delay of 100\,ms during unilateral operation, which is an acceptable margin for telesurgery  \citep{nankakuMaximumAcceptableCommunication2022}} and can be compensated for by the human brain \citep{miallAdaptationVisualFeedback2006, edenHapticCommunicationCentral2024}, and with no delay during bilateral teleoperation. These test conditions allowed us to directly test TIC and IAC during unilateral and bilateral teleoperation without additional confounds that may obscure comparisons. We carried out four separate experiments testing free space tracking, soft interaction with a balloon, insertion of a peg into a hole \citep{ajoudaniTeleimpedanceTeleoperationImpedance2012}, and bilateral teleoperation where a user modulated the follower's impedance whilst polishing a table while receiving force feedback. Supplementary Video\,1 exhibits TIC's and IAC's performance during free tracking and interaction with a soft object. IAC offers several advantages in comparison to state-of-the-art teleoperation methods \citep{Laghi2020} and related methods to achieve accurate follower control:
\begin{enumerate}
\item IAC achieves higher tracking accuracy than TIC given the same follower's impedance in free movement. This was verified in experiments for a follower's stiffness in the range of 80-1320\,N/m. 
\item IAC boasts superior tracking accuracy when the follower physically interacts with its environment.
\item IAC's high tracking accuracy enabled eight naive participants to complete a peg-in-hole task faster and more successfully relative to TIC.
\item IAC's stability is guaranteed through impedance shaping \citep{kronanderStabilityConsiderationsVariable2016} and is demonstrated during bilateral teleoperation where force feedback of the interaction between the follower and a hard surface was returned to the leader.
\end{enumerate}

\section{Problem statement}
\label{sec.problem}

\begin{figure}[tb]
\begin{center}
\includegraphics[width=0.95\linewidth]{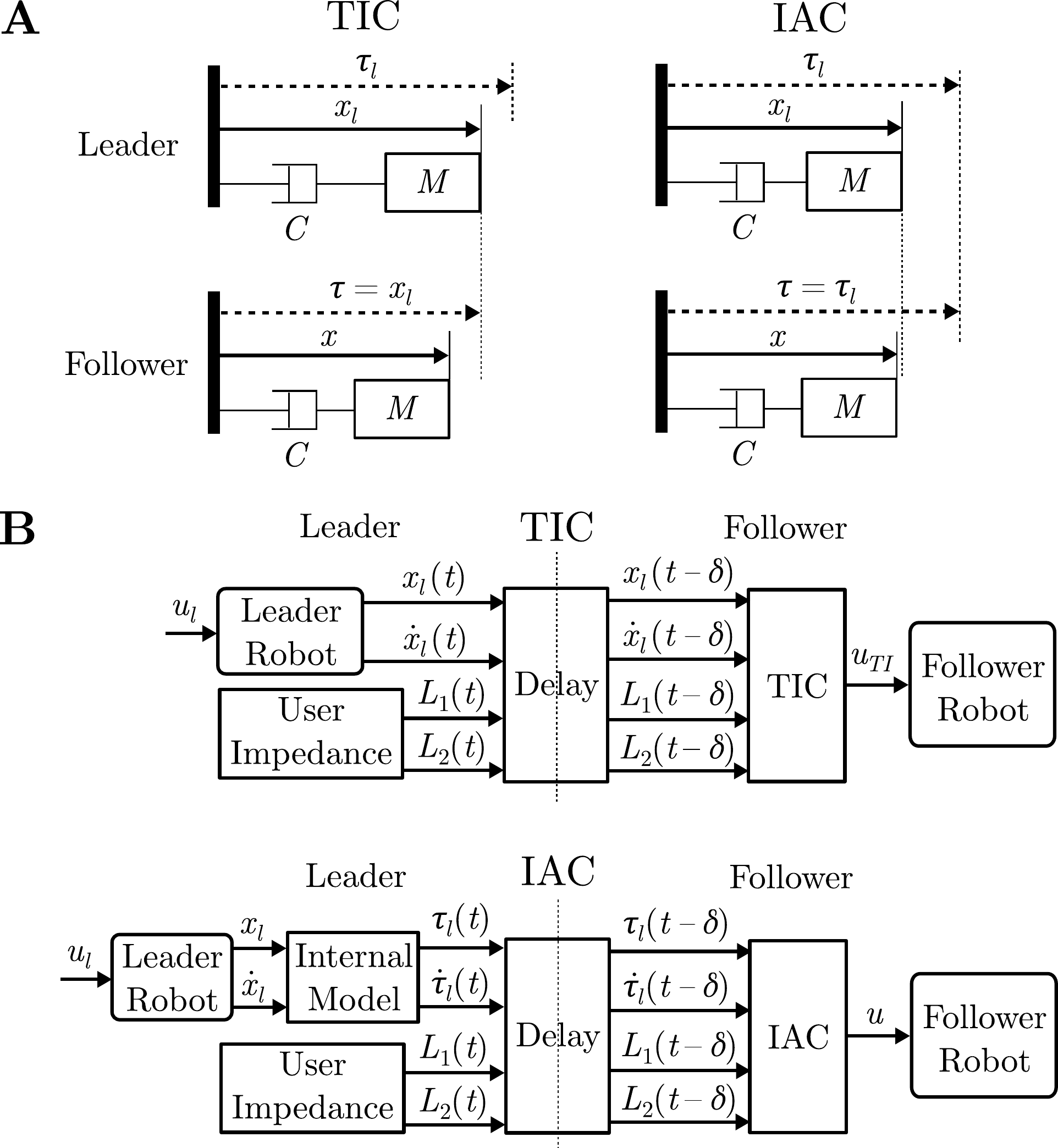}
\caption{Comparison of the novel intention assimilation control (IAC) with tele-impedance control (TIC). (A) A schematic illustration of the control for a follower robot tracking the leader's trajectory, where $x_l$ and $\tau_l$ are the leader's actual position and target position, and $x$, $\tau$ are the follower robot's actual position and target position, respectively. In TIC (left column), the leader's actual position is used as the follower robot's target, i.e. $\tau=x_l$. In IAC (right), the follower robot's target position is set as the leader's target position, i.e. $\tau=\tau_l$. In these schemes the gravity component is omitted for simplicity of illustration. (B) Block diagram of TIC (top) and IAC (bottom) used in all experiments. In TIC, the leader's position, velocity, and impedance parameters at time $t$ arrive at the follower after a delay period $\delta=100\,\textup{ms}$. In IAC, the leader's target position, velocity, and impedance parameters arrive with a delay.}
\label{fig:schematic}
\end{center}
\end{figure}

\subsection{System description}
We consider a teleoperation system with dynamics of the leader and the follower robots respectively given by
\bea
\label{eq.initial}
M(x_l)\ddot x_l + C(x_l,\dot{x}_l) \dot x_l + G(x_l) = u_l\\
M_f(x)\ddot x + C_f(x,\dot{x}) \dot x + G_f(x) = u_f
\eea
where $M(x_l)$, $C(x_l, \dot x_l)$ and $G(x_l)$ are the leader's inertia, Coriolis and centrifugal, and gravitational terms, respectively, and $M_f(x)$, $C_f(x, \dot x)$ and $G_f(x)$ those for the follower. $u_l$ and $u_f$ are the leader's and the follower's control commands, respectively. There are two possible interpretations of the leader's controller $u_l$: 1) for a passive leader robot, i.e., no torque/force from the robot's actuators, $M$, $C$, $G$ represent the original dynamic parameters of the leader robot; 2) for a leader robot whose dynamics have been already partially compensated for by the torque/force from the robot's actuators, $M$, $C$, $G$ represent the dynamic parameters of the robot after the compensation.

We consider the follower robot's controller
\bea
\label{eq.pre-compensation}
u_f \!\!\!\!&=&\!\!\!\! M_f(x)M^{-1}(x)(u-C(x,\dot{x})\dot x-G(x))\nb\\
&& +\, C_f(x,\dot{x})\dot x+G_f(x)
\eea
such that the follower robot's dynamics become
\bea
\label{eq.robot}
M(x)\ddot x + C(x,\dot{x}) \dot x + G(x) = u
\eea
where $u$ is the follower robot's new controller that will be designed below. The controller $u_f$ effectively compensates for the follower robot's original dynamics to make its new dynamics ``match'' the leader's dynamics.

\subsection{Problem formulation}
This paper studies a teleoperation scenario wherein the follower robot's controller is designed to make it follow the leader's trajectory. In particular, the leader's position $x_l$, velocity $\dot{x}_l$ and control command $u_l$ are measured subject to noise, and transferred to the follower robot.

Before introducing the IAC, we review the traditional method to design the follower robot's controller and analyze the issues to be addressed. As illustrated on the left-side of Fig.\,\ref{fig:schematic}A, a traditional follower robot's controller sets $x_l$ as its own reference trajectory. Without loss of generalisability, a feedback controller with gravity compensation for the follower robot can be designed as
\bea
\label{eq.robot-control-1}
u=G-L_1(x-x_l)-L_2(\dot x-\dot x_l)
\eea
where $L_1$ and $L_2$ are the follower robot's feedback controller gains or impedance parameters. Substituting this controller into Eq.\,\re{eq.robot} leads to
\bea
\label{eq.closed-robot}
M\ddot x + C\dot x +L_{2} (\dot x-\dot x_l)+ L_{1} (x-x_l)  = 0 \,.
\eea
When $L_1$, $L_2$ are positive definite, $C=0$, and $x_l$ is a constant, the above closed-loop dynamics will converge to $x=x_l$, i.e. the follower robot moves to the leader's target position. However, as $C$ is unlikely zero, e.g., due to friction, and $x_l$ is time-varying in a trajectory tracking task, the uncompensated dynamics will lead to a tracking error. One strategy to deal with these inevitable issues is to use large gains $L_1$ and $L_2$, but such high impedance is not suitable for a follower robot that could potentially collide with its environment.

Our goal here is to design a controller whereby the follower can accurately follow a leader's trajectory while freely choosing the follower's impedance to enable safe interaction during unexpected collisions. Our idea is to infer the operator's target position, and then design the follower robot's target position based on prescribed control gains. As illustrated in the right-side of Fig.\,\ref{fig:schematic}A, if the leader's target position $\tau_l$ can be accurately estimated, we can set $\tau=\tau_l$ and then a low-impedance control may suffice in ensuring the follower robot to track the leader. In particular, linearising the follower robot's model in Eq.\,\re{eq.robot} along the leader's trajectory $x_l$ \citep{OUYANG2006} yields
\bea
\label{eq.error}
\!\!\!\!\!\!\!\! && \!\!\!\! M(x_l)\ddot e + [C(x_l,\dot x_l)+S]\dot e + N e + O(\ddot e,\dot e,e,t) = u-u_l \nb \\
\!\!\!\!\!\!\!\! && \!\!\!\! S\!=\! \frac{\partial C}{\partial \dot x}|_{x_l,\dot x_l}\dot x_l, \, N \!=\! \frac{\partial M}{\partial x}|_{x_l}\ddot x_l+\frac{\partial C}{\partial x}|_{x_l,\dot x_l}\dot x_l+\frac{\partial G}{\partial x}|_{x_l}
\eea
where $e=x-x_l$ is the tracking error, and $O(\ddot e,\dot e,e,t)$ containing the higher-order terms of $\{\ddot e,\dot e,e,t\}$ is neglected.

Note that when $u=u_l$, Eq. \re{eq.error} can be considered as the leader's closed-loop dynamics, i.e.
\bea
\label{eq.leader-closed}
M(x_l)\ddot e+[C(x_l,\dot x_l)+S]\dot e+N e=0
\eea
which is stable and yields $e\rightarrow 0$. Therefore $S+C$ and $N$ are positive, which will be used when analyzing the follower robot's closed-loop dynamics. Moreover, Eq. \re{eq.leader-closed} can also be considered to be the follower robot's ideal closed-loop dynamics, which is achieved only when its control command $u$ is identical to the leader's $u_l$. Based on this idea, we explain how IAC is derived in the following section.

\section{Approach}
\label{sec.approach}
\subsection{Intention assimilation control}\label{sec.intention}
{Instead of transmitting the leader's current position $x_l$ to the follower as in Eq.\,\re{eq.robot-control-1}, we design the follower robot's controller by transmitting the leader's target position}, as
\bea
\label{eq.virtual}
u=-L_1(x-\tau^*)-L_2(\dot x-\dot \tau^*)
\eea
The above controller has the same form as Eq.\,\re{eq.robot-control-1} but with a virtual target $\tau^*$ replacing $x_l$. In the following, we will demonstrate that the follower's impedance parameters $L_1$ and $L_2$ do not necessarily have to be large as in Eq.\,\re{eq.robot-control-1} for accurate tracking. Therefore $L_1$ and $L_2$ can be set by the designer according to the context of a task, e.g. low if the robot is to move through a cluttered environment during unwarranted collisions, or high if the follower is expected to penetrate into hard materials like during drilling. Importantly, the gains $L_1,L_2$ can also be adapted online to ensure system stability {\citep{kronanderStabilityConsiderationsVariable2016}}, or modulated according to the human operator's impedance \citep{Laghi2020}.

Since the leader's and follower's dynamics in Eqs.~\re{eq.initial} and \re{eq.robot} are identical, the follower robot's movement would be identical to the leader's if their controllers are the same. Thus, $u=u_l$ can be expanded to
\bea
\label{eq.equal}
-L_1(x-\tau^*)-L_2(\dot x-\dot \tau^*)=u_l\,.
\eea
The follower robot's target $\tau^*$ can be computed from the above equation if $x$, $\dot x$, $u_l$ are measurable, as explained in the following. For convenience of analysis, we consider the proposed approach in a single dimension of the task space.

In order to estimate $\tau^*$, we consider the internal model
\bea
\label{eq:targetorder}
\tau^*=\theta^T\phi\,, \quad \phi=[1, t,\ldots,t^n]^T
\eea
where $\theta^T$ is the transpose of the parameter vector determining $\tau^*$ and $n$ the order of the target model $\tau^*$. For example, $\tau^*=p$ indicates a fixed position while $\tau^*=p+v t$ is a trajectory starting from position $p$ and moving with velocity $v$. $x$, $\dot x$, $\theta$ and $u_l$ are concatenated as the state vector $\xi \equiv [x,\dot x,\theta^T,u_l]^T$, to which the above internal model is extended to
\bea
\label{eq.state-space}
\dot\xi= \!\left[
                 \begin{array}{c}
                   \dot x \\
                   \ddot x \\
                   0 \\
                   \dot u_l \\
                 \end{array}
               \!\right]\!+\nu
\eea
where $\nu \in N(0,\sigma_s)$ is system noise to identify $\theta$. 
The assumed measurement yields the observation equation
\be
\label{eq.measurement}
z\equiv \left[
                 \begin{array}{c}
                   x \\
                   \dot x \\
                   u_l \\
                 \end{array}
               \!\right]\!+\mu
\equiv H\xi+\mu \,, \,\,\,
H\equiv\left[\!\!
\begin{array}{cccc}
1 & 0 & 0 & 0\\
0 & 1 & 0 & 0\\
0 & 0 & 0 & 1\\
\end{array}
\!\!\right]
\ee
where $\mu \in N(0,\sigma_m)$ is measurement noise. 

The following system observer is developed to compute the robot's estimate of the extended state
\bea
\label{eq.observer}
\dot{\hat\xi}\!\!\!\!&=&\!\!\!\left[\!\!
\begin{array}{c}
\dot{\hat x} \\
\ddot{\hat x} \\
0 \\
\dot{\hat u}_l \\
\end{array}\!\!
\right] + \, K(z \! - \!\hat {z}) \nb \\
\hat u_l \!\! &=& \!\!-L_{1}\! (x\!-\!\tau)- L_{2}(\dot x-\dot\tau)\nb\\
\hat {z} \!\! &=& \!\! H \hat \xi,~\tau=\hat \theta^T\phi
\eea
where $\hat{\cdot}$ represents the estimate of the corresponding variable minimizing $E(\|\hat{\xi}-\xi\|^2)$ and $\tau$ is the estimate of $\tau^*$, defined as the leader's \textit{virtual target}. $K$ is the Kalman gain, which is updated according to
\bea
\label{eq.K}
K=P H^TR^{-1}
\eea
where $P$ is a positive definite matrix obtained by solving the Riccati equation
\bea
\label{eq.riccati}
P A^T+ A P-P H^TR^{-1} HP+Q=\dot P \, ,
\eea
with the noise covariance matrices $Q \equiv E[\nu \nu^T]$ and $R \equiv E[\mu \mu^T]$. $A$ is the system matrix obtained by linearizing Eq.\,\re{eq.state-space} to the following form
\bea
\label{eq.state-space-linear}
\dot\xi&=& \, A \xi + B u+ \, \nu,
\quad B = \left[\!\!\!
\begin{array}{cccc}
0\\
M^{-1}\!\!\!\\
0\\
L_2 M^{-1}
\end{array}
\!\!\!\right]\!, \nb\\
A\!\!\!\!\!&=&\!\!\!\!\!\left[\!\!
\begin{array}{cccc}
0 \!&\! 1 \!&\! 0 \!&\! 0\\
0 \!&\! -M^{-1}C \!&\! 0 \!&\! M^{-1}\\
0 \!&\! 0 \!&\! 0 \!&\! 0\\
0 \!&\! -L_1\!+L_2 M^{-1}C \!&\! L_1\dot\phi \!&\! -L_2 M^{-1} \\
\end{array}
\!\!\!\!\right]
\eea
where the second row of $A$ is from Eq.\,\re{eq.robot}, and the last row is obtained by differentiating Eq.\,\re{eq.equal} and substituting Eqs.\,\re{eq.robot} and \re{eq:targetorder}.

The estimate of the observable system state $\hat{\xi}$ yields $\hat u_l=u_l$. Therefore, if the follower robot is moving in free space, it can track the leader's movement with the controller (as proved in the Appendix):
\bea
\label{eq.fm}
\hat u_l=-L_1(x-\tau)-L_2(\dot x-\dot {\tau}) \,.
\eea
$L_1$ and $L_2$ can take on any arbitrary positive values, unlike the high-impedance controller in Eq.\,\re{eq.robot-control-1}. In the following two subsections, we explain how $L_1$ and $L_2$ can be prescribed by the operator and then how the intention assimilation control is implemented.

\begin{figure}[tb]
\begin{center}
\includegraphics[width=1\linewidth]{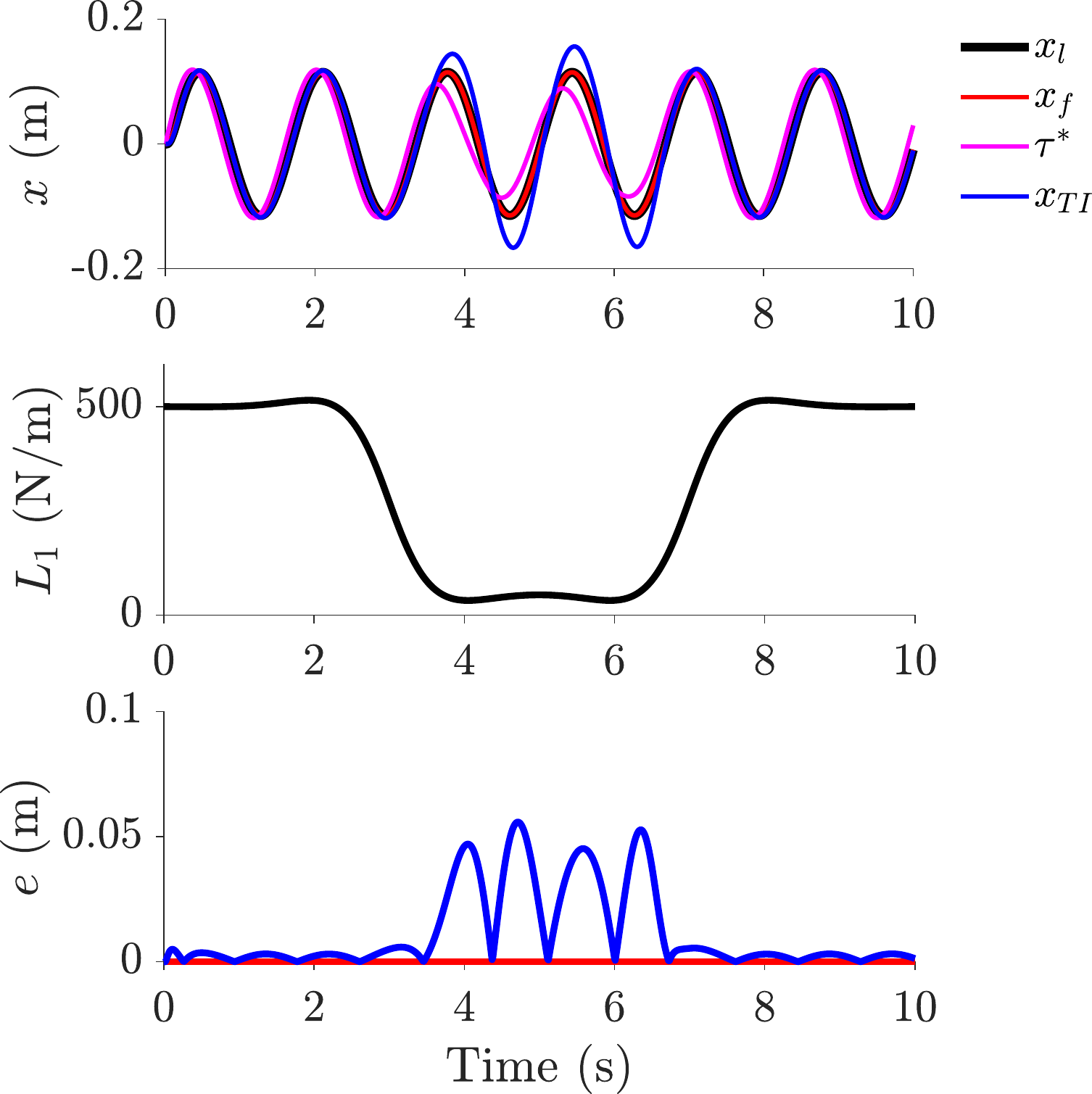}
\caption{Simulation of a point mass follower under TIC ($x_{TI}$, in blue) and IAC ($x_{f}$, in red). Unlike TIC, IAC maintains a small positional error $e$ between the leader and follower even when the follower's impedance is lowered by the operator in real-time.}
\label{fig:gainchange}
\end{center}
\end{figure}

\subsection{Tele-impedance with accurate tracking}
\textit{Tele-impedance control} (TIC) was introduced as a technique to transfer human impedance to the follower robot, using EMG to estimate human arm impedance \citep{Laghi2020}. {With TIC the follower robot is commanded to move through the control input
\bea
\label{eq.uTI}
u_{TI}=-L_{1}(x-x_l)-L_2(x-\dot{x}_l).
\eea
$L_1$ and $L_2$ are the user's endpoint stiffness and damping, which are estimated from muscle activity that must be calibrated beforehand.}

Alternatively, we have recently shown that the arm impedance magnitude can be simply estimated as it linearly increases with the grasp force \citep{takagiEndpointStiffnessMagnitude2020}. In particular, we expand the leader's control as
\bea
u_l=-L_{l1}(x-\tau_l)-L_{l2}(\dot x-\dot \tau_l)
\eea
where $L_{l1}$ and $L_{l2}$ are human arm impedance parameters. If they can be accurately estimated, we can set $L_1 \equiv L_{l1}$ and $L_2 \equiv L_{l2}$ so that the follower robot mirrors the human's impedance. 

These parameters can be used in Eq.\,\re{eq.robot-control-1}, yielding TIC. They can be also used in the proposed IAC, in which case we have $\tau^* = \tau_l$ from Eq.\,\re{eq.equal}, i.e. the follower robot's target position is identical to the leader's. As a result, IAC transfers the leader's impedance to the follower robot while accurately tracking the leader's movement. Interestingly, accurate tracking of the leader's movement is still guaranteed even if the estimation of human arm impedance is inaccurate, as will be shown in the experimental results. As discussed in the previous subsection, this is because tracking of the leader's movement is independent of prescribing the follower robot's impedance, i.e. $\tau^*$ is calculated using Eq.\,\re{eq.equal} based on the follower robot's impedance that ensures $u=u_l$. Due to this feature, other impedance adaptation approaches can be integrated with IAC and the corresponding stability analysis could be carried out.

{In the event of a communication delay $\delta$ from the leader to the follower robot, TIC's control input at time $t$ is
\begin{eqnarray}
u_{TI}(t)\!&\!\!\!\!=\!\!\!\!&\!-L_1(t_\delta)[x(t)\!-\!x_l(t_\delta)] \!- \!L_2(t_\delta)[\dot{x}(t)\!-\!\dot{x}_l(t_\delta)]\,, \nonumber \\
t_\delta \! &\!\!\!\!\equiv\!\!\!\!& t \!-\! \delta
\label{eq.uTIdelay}
\end{eqnarray}
and for IAC it is
\begin{equation}
u(t)\!= \!-L_1(t_\delta)[x(t)\!-\!\tau^*\!(t_\delta)] \!- \!L_2(t_\delta)[\dot{x}(t)\!-\!\dot{\tau}^*\!(t_\delta)]\,.
\label{eq.uIACdelay}
\end{equation}
}

\subsection{Stability when adapting the follower's impedance}

TIC or IAC will exhibit stable behavior if the follower's impedance remains constant during operation. However, they can become unstable if the follower's damping is small and its stiffness is increased too rapidly. To ensure stable operation, we limit the rate of leader stiffness increase, as proposed in \citep{kronanderStabilityConsiderationsVariable2016}, before transferring it to the follower. Setting
\bea
\label{eq.alpha}
\alpha = \min \frac{\underline{\lambda}(L_2)}{\overline{\lambda}(M)}
\eea
with $\underline{\lambda}(\cdot )$ and $\overline{\lambda}(\cdot )$ the smallest and largest eigenvalues, respectively, the change rate of $L_1$ is constrained to satisfy
\bea
\label{eq.gainrestriction}
\dot{L}_1<2\alpha L_1 - \alpha \dot{L}_2 \,\,.
\eea

\subsection{Bilateral teleoperation}

To compare the performance of TIC and IAC during bilateral teleoperation, we set the communication delay to 0 ms to avoid unnecessary confounds that may arise from the inclusion of passivity algorithms. Without communication delay, the interaction force measured by the follower robot $F_{env}$ can be transmitted directly to the leader robot so that it can perceive the force experienced by the follower. Since IAC uses the leader's force to estimate their target, the interaction force must be subtracted from the total force $u_t$ to accurately estimate the force that the user exerts on the leader robot. Thus, the observation equation Eq.~(\ref{eq.measurement}) is modified to 
\be
z\equiv \left[
                 \begin{array}{c}
                   x \\
                   \dot x \\
                   u_t - F_{env} \\
                 \end{array}
               \!\right]\!+\mu
\ee
during bilateral teleoperation.

\section{How intention assimilation control works}
\label{sec.simulation}
A simulation with a leader and a follower robot moving along a one-dimensional axis is used to illustrate the functioning of IAC, and compare it to the state-of-the-art TIC of \citep{Laghi2020}. For this simulation we assume that $L_1$ and $L_2$ are known at all times, and neglect sensorimotor and measurement noises. The leader moves along a sinusoidal trajectory with an amplitude of 10\,cm and a frequency of 0.6\,Hz (Fig.\,\ref{fig:gainchange}A). The follower tracks the leader's movement at the start of the simulation using the impedance $L_1$\,=\,500\,N/m. From 2 to 8 seconds into the simulation, $L_1$ is rapidly lowered to 50\,N/m.

We see in Fig.\,\ref{fig:gainchange} that TIC has trouble ensuring accurate tracking of the leader's trajectory, where the follower overshoots the leader's position when the follower's impedance is too low. IAC, on the other hand, modifies the desired trajectory $\tau^*$ in accordance with the change in the follower's impedance (magenta trace in Fig.\,\ref{fig:gainchange}). As a consequence the follower's tracking error remains unchanged despite its impedance being lowered to a tenth of its original value at the beginning of the simulation (Fig.\,\ref{fig:gainchange} bottom panel, red). 

In summary, TIC has difficulty maintaining high tracking accuracy when the follower's impedance is low, and IAC can overcome this limitation. In practice, the human operator could cocontract the muscles in their arm, thereby increasing the follower's impedance to accurately track their movement with TIC. However, cocontracting muscles {requires large effort} and cannot be sustained for a long period. Alternatively, the TIC designer could constrain the follower's impedance such that it remains stiff even when the user relaxes. However, this would restrict the range of impedance that the operator can control, thereby undermining TIC's advantage of enabling the user to control the follower's impedance in a feedforward manner.

\section{{Experiments}}

\begin{figure}[tb]
\begin{center}
\includegraphics[width=1\linewidth]{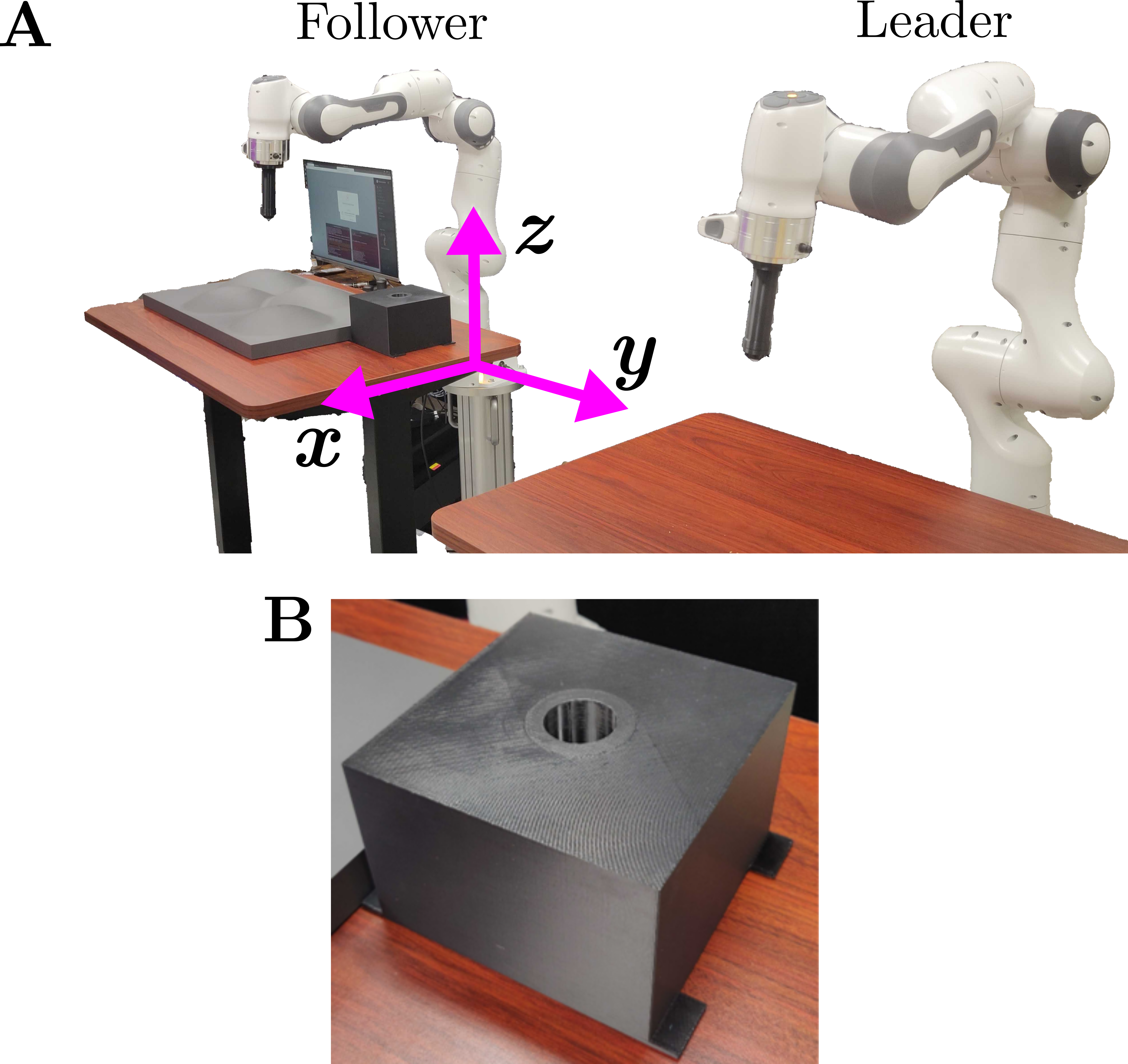}
\caption{Experimental setup. (A) Photo of the leader and follower robotic arms used in the experiments. (B) 3D printed box with a cylindrical hole used for the peg-in-hole experiment.}
\label{fig:setup}
\end{center}
\end{figure}

\begin{figure*}[hbt]
\begin{center}
\includegraphics[width=1\linewidth]{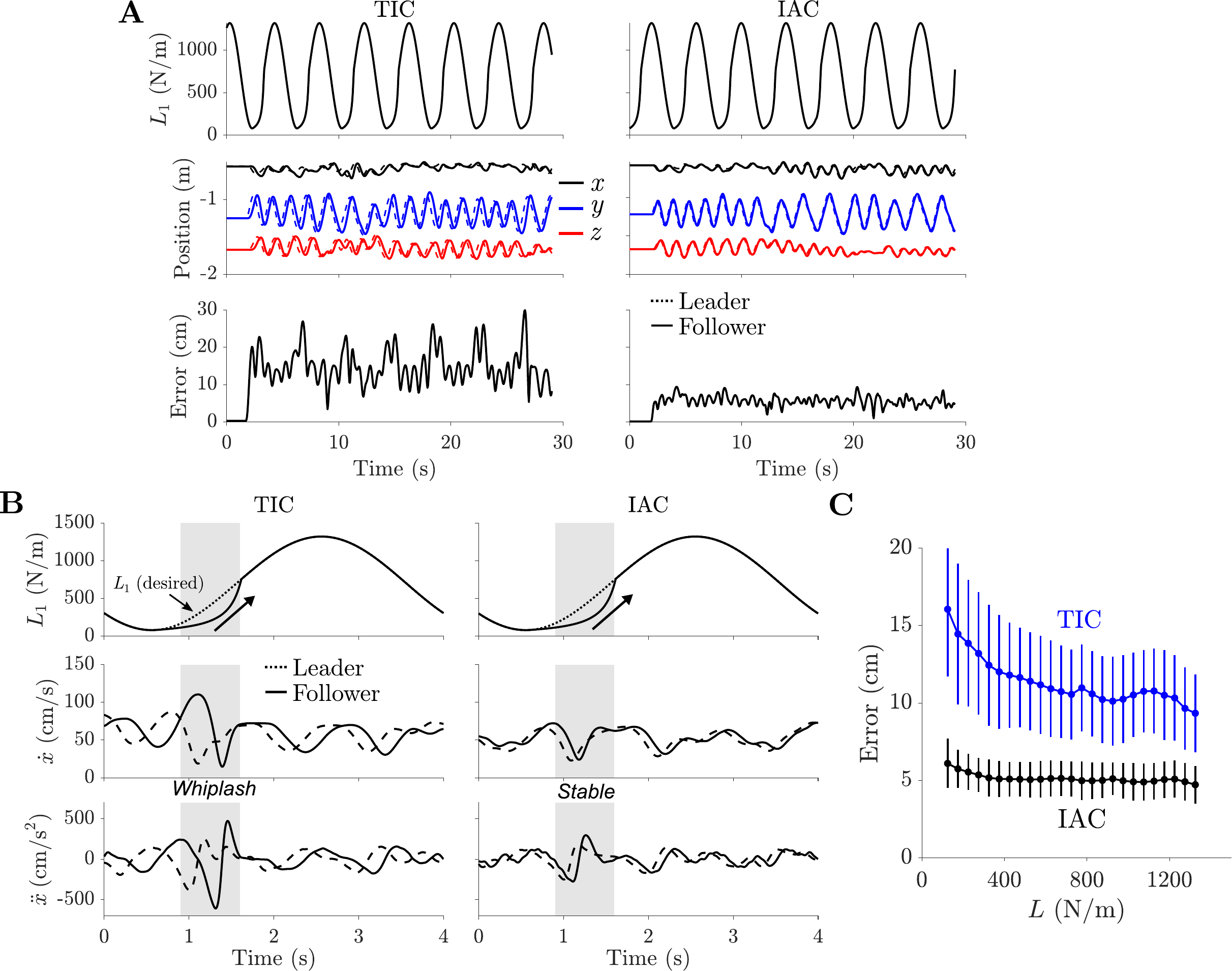}
\caption{Comparison of teleoperation performance during free tracking by an operator. (A) Free tracking error using TIC (left column) vs. IAC (right). The follower's impedance was updated automatically while the leader was moved by the operator along all three Cartesian axes. For clarity, only a 30\,s segment is shown. The tracking error, computed as the Euclidean distance between the leader and follower trajectories, is low with IAC irrespective of the impedance settings. (B) When the follower's impedance increased slowly from 80\,N/m, the increase in the desired impedance was stabilized with both TIC and IAC to ensure stable operation (dashed lines in top panels are the desired stiffness, and solid lines are the stabilized stiffness). With TIC,
the follower exhibited whiplash wherein its acceleration rapidly increased and decreased to hone in and track the leader's trajectory, but there was no whiplash with IAC. (C) Mean tracking error as a function of the follower's impedance during free tracking. Error bars represent standard deviation. With TIC, error increased sharply when the follower's impedance was below 400\,N/m, whereas IAC maintained the follower close to the leader's trajectory even with low impedance.}
\label{fig:freetracking}
\end{center}
\end{figure*}

We validated IAC in experiments with a setup using two robotic manipulators (Franka Emika, Germany). These 7 degrees-of-freedom robots are equipped with torque sensors at each joint, enabling it to estimate external torques and forces. On top of the pre-implemented gravity compensation controller, torque commands can be sent to each joint to control its impedance. The robots were placed side-by-side with a user controlling the follower robot by moving the leader robot (Fig.\,\ref{fig:setup}). In all other experiments besides the bilateral teleoperation experiment, a {software-imposed} constant communication delay of $\delta=100\,\textup{ms}$ was set  when sending control parameters from the leader to the follower robot. {Users viewed the follower robot directly while interacting with the leader robot. Both the leader and follower robots were controlled at 1000\,Hz, and the data was recorded at this rate too.}

Four experiments were conducted to compare and contrast the performance of IAC versus TIC: a free tracking experiment, an experiment testing interaction with a soft object, a peg-in-hole experiment, and a bilateral polishing experiment. The peg-in-hole study involved eight naive participants, and all other experiments were carried out by one operator. Three types of information were sent from the leader to the follower: the target position, the target velocity, and the impedance along the $x$, $y$, and $z$ axes. The stability condition Eq.\,\re{eq.gainrestriction} was applied on the leader's desired impedance with $M$=\,12.8\,kg (moving mass of the follower robot). For IAC, the estimated target trajectory was calculated using the stabilized impedance values. The studies were approved by the institution's research ethics committee (no. R05-014), and all participants provided written informed consent prior to participation. In the following experiments' description the $x$ and $y$ Cartesian axes are in the table plane, and the $z$ axis is perpendicular to it. In all experiments, the damping parameter is related to stiffness with $L_2$ = 0.1$L_1$. Eq.\,\re{eq.fm} was solved directly in the experiments as the measurement of $u_l$ was relatively accurate on the leader robot. Furthermore, the implementation of the Kalman filter in Eqs.\,\re{eq.observer}-\re{eq.riccati}, which was carried out in the simulation, reduced the control rate by a factor of ten when implemented on all seven joints. The rotational stiffness was set to 15\,Nm/rad unless specified otherwise.

\textit{The free tracking experiment tested the follower's ability to accurately track the leader's movement in the free space}. The experimenter displaced the leader's handle in continuous movements with a magnitude spanning 20-40\,cm along each axis. The movements were usually across two or three axes simultaneously (see Fig.\,\ref{fig:freetracking}). During these movements, the follower's impedance along the $x, y, z$ axes was changed automatically by the control system according to {$L_1 = 700 + 620\sin(0.25\pi t)$\,N/m, }such that the follower's impedance was modulated between {80-1320\,N/m} with a period of 4\,s. Two 60 seconds trials were conducted, one where the follower was controlled using TIC, and the other with IAC.

\begin{figure*}[h]
\begin{center}
\includegraphics[width=1\linewidth]{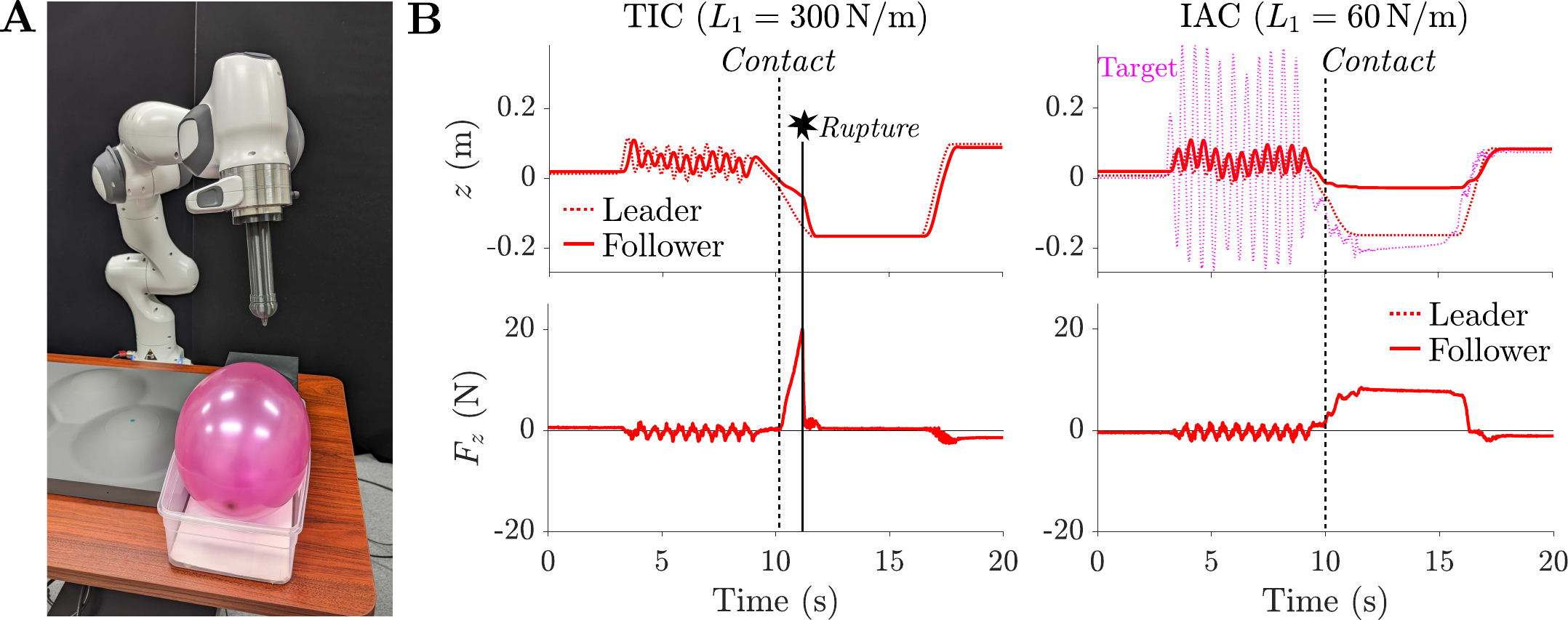}
\caption{{Balloon experiment showcasing the advantage of a follower that is compliant yet accurately tracks the leader's motion (Supplementary Video\,1). (A) Photo of the robotic arm equipped with a 3D printed tip and the balloon. (B) An operator made rapid vertical motions along the $z$ axis to make the follower move up and down above the balloon for 5 seconds, and then moved straight down until the operator touched the leader's table whose height was slightly higher than the follower's. The follower was controlled using either TIC with a gain of $L_1$=300\,N/m or using IAC with a gain $L_1$=60\,N/m. Despite the large gain used in TIC, the follower did not accurately track the leader's position in free space, and it ruptured the balloon as the leader moved downward. With IAC, the follower tracked the leader's motion by following the estimated target position $\tau^*$ (magenta) and compliantly interacted with the balloon so as not to break it.}}
\label{fig:balloon}
\end{center}
\end{figure*}

The second experiment tested interaction with a soft object. The experimenter moved the leader robot along the $z$ axis for 5 seconds above the balloon's position, then moved downward rapidly until they made contact with the leader's table whose height was positioned 5\,cm higher than the follower's (Fig.\,\ref{fig:balloon}A). IAC's gain was $L_1$\,=\,60\,N/m for this experiment while for TIC's $L_1$\,=\,300\,N/m as the follower could not keep up with the leader's motion when the gain was smaller. Despite this five-fold difference in the gain, the tracking performance with IAC was superior to TIC (refer to Fig.\,\ref{fig:freetracking}C).

\textit{A third experiment requiring peg-in-hole control} was carried out by eight participants (aged 35$\pm$2, two females) who had no previous experience in controlling the robot. Participants were instructed to insert a cylindrical peg attached to the follower robot into the hole of a box, which was attached to a table using strong adhesives as shown in Fig.\,\ref{fig:setup}B. The cylindrical peg was 10.7\,cm long with a diameter of 32\,mm. The hole had a length of 10.2\,cm and a diameter of 33\,mm. Both the peg and the hole were 3D printed using ABS as the printing material. The follower's starting position was at [0.04, 0.07, -0.08]\,m and the hole's entry position at [-0.12, 0.29, -0.06]\,m.

For safety, participants were first given time to familiarize themselves with the leader robot by moving it in free space. They then completed 10 trials in total, trials \{1,3,5,7,9\} during which the follower was controlled using IAC, and trials \{2,4,6,8,10\} by TIC. The trials were staggered to evenly distribute the motor learning of the task between the two conditions. The follower's stiffness was fixed to 60\,N/m in the $x$ direction, 40\,N/m in $y$, 300\,N/m in $z$ to directly compare the task performance with TIC and IAC without any additional confounds. A trial was successful when the peg was inserted all the way into the bottom of the hole within 20 seconds. If a trial lasted more than 20\,s, it was stopped and considered a failure. The rotational stiffness along all axes was set to 4\,Nm/rad for the peg-in-hole task.

A fourth \textit{bilateral polishing experiment} was then conducted to test the follower's performance when its impedance was modulated by the leader via their power grasp force, which was measured using a USX10-H10 force sensor (Tec Gihan, Japan). {A table was placed underneath the leader and follower robots with the leader's table positioned approximately 4\,cm below the follower's. The user moved along the edge of a square of length 10\,cm along the $x$-$y$ plane whilst modulating their grasp force so that the follower robot exerted a varying force onto the table. Two 30\,s trials were conducted, one with TIC and the other with IAC. The leader received force feedback of the interaction between the follower and its table.

\section{Experimental results}
\subsection{Free tracking}
In the free tracking experiment the operator, who was familiar with the robots, made multi-axis movements with the leader robot while the follower's impedance smoothly oscillated between highly compliant and stiff settings. Fig.\,\ref{fig:freetracking}A shows the resulting follower's impedance, which was identical across all three Cartesian axes, the leader and follower robots' position and their difference as a function of time. Here the ``error'' is computed as the Euclidean distance of the differences in the $x$, $y$, and $z$ coordinates.

When the follower was controlled with TIC, the error jumped when the follower's impedance approached the minimum value of 80\,N/m. Each time the error jumped it exceeded a distance of 20\,cm. Even when the follower's impedance was high, the error hovered around 12\,cm. However, the follower's tracking performance was significantly improved when using IAC. While the follower's error increased when its impedance approached its most compliant value of 80\,N/m, the error never exceeded 10\,cm and hovered around 5\,cm. Supplementary Video\,1 shows TIC versus IAC during free tracking when the impedance along all axes was fixed to 100\,N/m. On this video, it is clearly visible that the follower robot does not accurately track the leader's movement with TIC, but with IAC the tracking is more accurate.

The large jumps in error occurred when the follower's impedance was most compliant. After the follower's impedance was increased gradually from 80\,N/m with TIC, the follower exhibited whiplash wherein it accelerated and decelerated rapidly to track the leader's motion (Fig.\,\ref{fig:freetracking}B). Such a whiplash was avoided with IAC. As whiplash is highly undesirable in teleoperation settings, this greatly limits the range and the rate at which the follower's impedance can be changed by the user.

We further assessed the performance of the two controllers during the free tracking task by binning the error as a function of the follower's impedance, which is charted in Fig.\,\ref{fig:freetracking}C. The follower's error and its deviation were consistently higher for all impedance values with TIC. Relative to IAC, the follower's error was 2.21 times greater. Even when restricting the comparison to when the follower's impedance was greater than 1000\,N/m, the follower's error was still 2.08 times greater with TIC relative to IAC. It appears that the tracking performance was significantly improved with IAC for the same impedance across a wide range of values between 80\,N/m and 1320\,N/m.

{Balloon interaction}
Fig.\,\ref{fig:balloon}B shows the results from the experiment comparing a follower controlled using TIC and IAC when interacting with a balloon. The leader moved up and down rapidly in free space to illustrate the tracking performance of each controller, and then was moved downward towards the leader's table so that the follower made contact with the balloon on the follower's table. Despite TIC having a higher gain than IAC (300\,N/m versus 60\,N/m), IAC's free tracking performance was superior. Under TIC, the force measured by the follower increased rapidly as it pushed into the balloon, eventually rupturing it. With IAC the follower gently pressed against the balloon and did not rupture it. This showcases the advantage of having a compliant gain in a remote environment replete with soft objects (see Supplementary Video\,1), demonstrating IAC's ability to keep interactions safe at low impedance levels whilst simultaneously keeping the tracking accuracy high in free space.

\begin{figure*}[h]
\begin{center}
\includegraphics[width=1\linewidth]{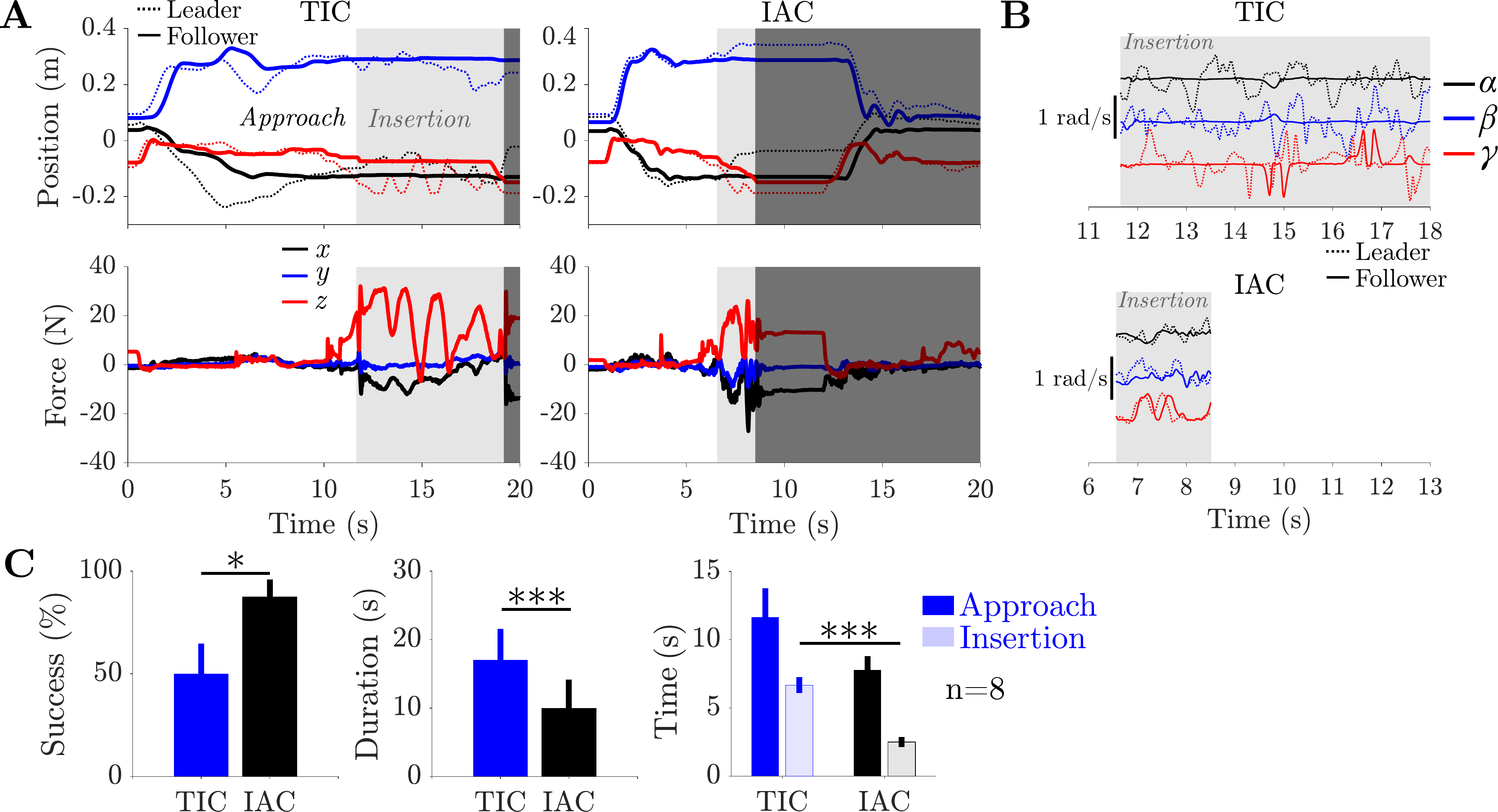}
\caption{Peg-in-hole performance using TIC and IAC. Panels A and B illustrate the typical performance of an exemplar participant, while C analyzes performance across eight participants. (A) Sample trials from a peg-in-hole task using TIC (left column) and IAC (right) from the same exemplar participant. The background color indicates the approach phase (white), the insertion phase (grey), and the task completion phase (black). Under TIC, the follower was unable to track the leader's movement along the $x$ and $y$ axes. With IAC, the follower rapidly followed the leader's movement, and the participant quickly inserted the peg into the hole. (B) Orientation time-series of the leader and follower robots from the same trials during only the insertion phase. The participant tried to adjust the follower robot's orientation during insertion to get it unstuck, but the follower did not respond to the leader's rotational motion under tele-impedance. With IAC, the follower's orientation could be adjusted and corrected to jiggle it into the hole. (C) Success rate, duration, and time spent during approach and insertion in the peg in hole experiment with TIC and IAC from all eight participants. Success rate was greater with IAC and when analyzing only successful trials, the task duration was also significantly shorter. The time spent during the approach phase was comparable between TIC and IAC, but the insertion was much faster with the latter. * indicates $p<0.05$ and *** $p<0.001$.}
\label{fig:peginhole_sampletrial}
\end{center}
\end{figure*}


\subsection{Peg-in-hole}
Fig.\,\ref{fig:peginhole_sampletrial}A shows two sample trials (last trial of each condition) of the peg-in-hole task completed by the same exemplar participant using TIC and IAC. In the approach phase, the participant moved the follower robot mainly along the $x$ and $y$ axes to transport the peg towards the hole and align it for insertion. This was followed by the insertion phase, which began when the peg's tip was inside the hole, and ended when the peg reached the end of the hole. The insertion axis was aligned with the $z$ axis.

With TIC, the participant had difficulty transporting and aligning the peg with the hole, which is discernible from the time taken to complete the approach phase. And even in the insertion phase, where the peg was already partially inside the hole at the beginning, the participant struggled to lower the follower's $z$ position to push the peg further into the hole. The participant moved the leader's $z$ position up and down to try and move the follower further into the hole, but only did so successfully after several attempts and a significant amount of time. With IAC, the follower robot accurately tracked the leader's motion during transport, enabling the participant to rapidly align the follower robot to the hole's position. 

Importantly, we also observed a reduction in the time taken for insertion. This improvement may seem surprising as the approach phase benefits from superior free tracking performance, but that is not the case during interaction when the peg is inside the hole. So why did participants complete the insertion phase faster with IAC relative to tele-impedance, despite the fact that the robot's impedance was identical in both TIC and AIC conditions?

During insertion, the peg's orientation plays a vital role in determining whether it smoothly enters the hole or gets stuck somewhere inside. We compared the follower's three-axis angular velocity with respect to the leader's one during the insertion phase (Fig.\,\ref{fig:peginhole_sampletrial}B). With TIC the participant continuously changed the leader's orientation in an attempt to dislodge the follower's peg, which was stuck somewhere in the hole, without success as the follower's orientation did not respond to the leader's change in orientation. With IAC, the follower's angular velocity along all axes did respond to the leader's change in orientation, which made it easier to dislodge the follower's peg when it got stuck inside the hole. Thus, the ability to dislodge the peg by changing the follower's orientation was likely key to improving insertion time and to a higher task completion rate.

These observations are confirmed by the data of eight naive participants, where we analyzed the quantitative difference in task performance using the two controllers. First, we looked at how many trials participants succeeded at completing the peg-in-hole task within the allocated time of 20 seconds. Fig.\,\ref{fig:peginhole_sampletrial}C shows a summary of the results. With TIC, participants finished the task within the allocated time on 50$\pm$15\% of trials. With IAC, participants succeeded on 88$\pm$8\% of trials, a significant improvement over TIC (t(7)=2.9,\,p=0.02). We then analyzed how long it took participants to complete the task. For this analysis, only successful trials were used. In total, it took 17.0$\pm$1.6\,s with TIC and and 10.0$\pm$1.5s with IAC, a reduction that proved significant (t(7)=-9.6,\,p$<$0.001). While the time taken during the approach phase contributed to the reduction in trial time, most of the decrease in trial duration came from the insertion phase, which was much faster with IAC (2.5$\pm$0.4\,s versus 6.7$\pm$0.6\,s, t(7)=-8.4,\,p$<$0.001).



\subsection{Bilateral polishing}
Fig.\,\ref{fig:bilateral_tablepolish} shows the results of the bilateral table polishing experiment. When the follower's impedance was increased by the operator's grasp force while maintaining a roughly constant position beneath the table's surface, the follower robot exerted an increasingly large force against the table to track the operator's motion as the mechanical impedance increased. With TIC, an increase in the follower's impedance resulted in a reduction in the tracking error on the $x$-$y$ plane. On the other hand, a small follower impedance made the tracking error jump in value as the follower did not follow the leader's movement. This was avoided by using IAC, which enabled the follower to accurately track the leader's movement along the plane of the table. Furthermore, as the operator raised their grasp force to increase the follower's impedance, the polishing force against the table grew larger with the commanded impedance. The rise and fall in the polishing force against the table was reflected back to the operator in both TIC and IAC, showing accurate reflection of the follower's interaction force and stable bilateral teleoperation with both control modalities.

\begin{figure*}[h]
\begin{center}
\includegraphics[width=0.8\linewidth]{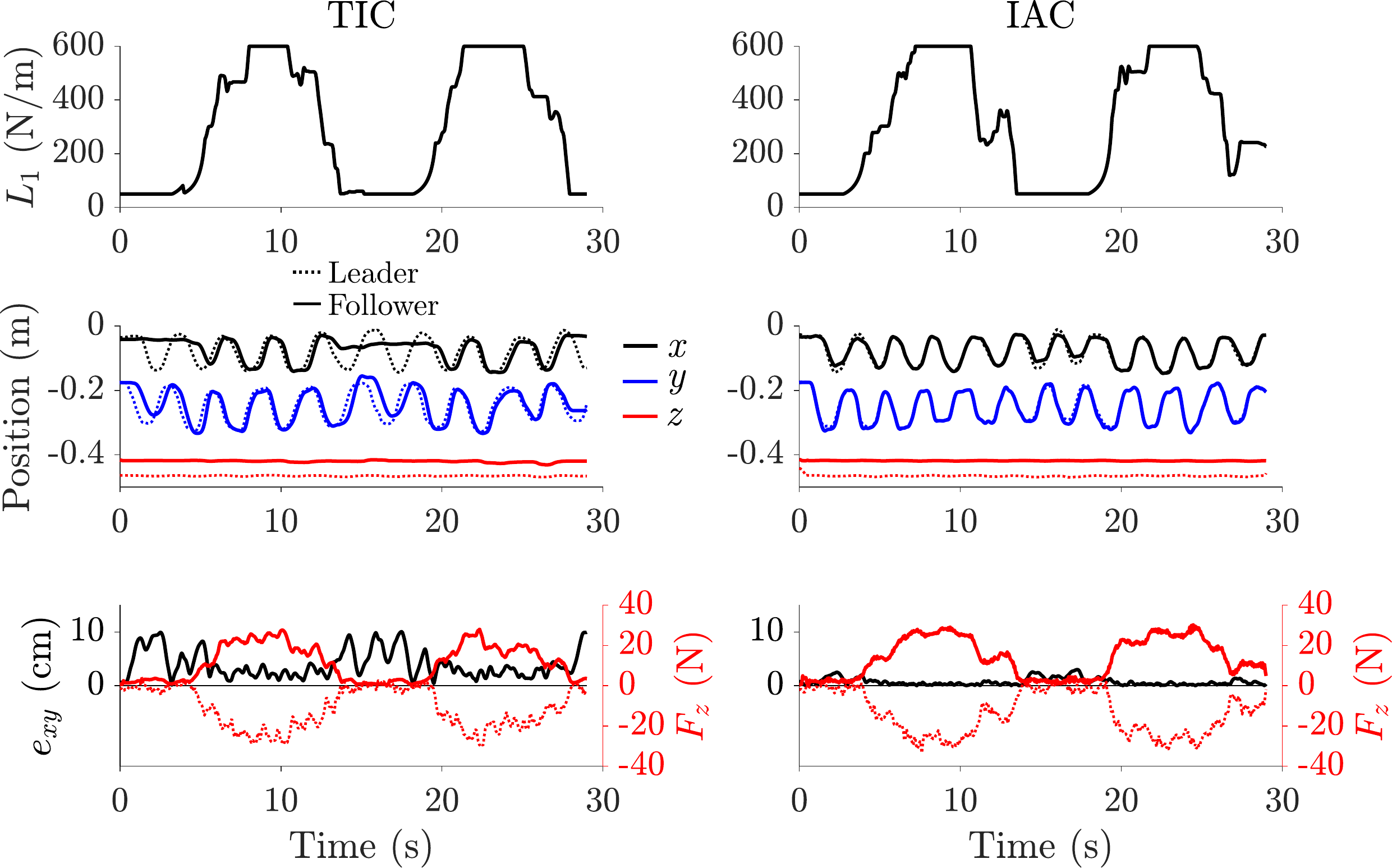}
\caption{Results of the bilateral polishing experiment, where an operator traced a square with the leader robot, ensuring that the follower robot polished the table normal to the $z$ axis, while the follower's impedance was modulated by the leader's power grasp force. The error $e_{xy}$, defined as the Euclidean distance between the leader and follower along the free $x$ and $y$ axes, increased significantly with TIC when the follower's impedance was small, but remained consistently small with IAC. A noticeable increase in impedance was observed through the large force $F_z$ measured by the follower robot as it pushed against the table when the follower's impedance was increased by the operator's grasp force. The operator perceived the equal but opposite force feedback of the follower's interaction with the table.}
\label{fig:bilateral_tablepolish}
\end{center}
\end{figure*}

\section{Discussion}
In both unilateral and bilateral teleoperation, it is desirable for the follower's impedance to match the leader's impedance, but with communication delays this becomes challenging. TIC was proposed as a solution to this problem as it enables the operator to directly control the follower's impedance. However, TIC's tracking accuracy degrades rapidly as the follower's impedance is made compliant, which becomes evident for compliance values smaller than $<$\,600\,N/m. To address this problem by proposing IAC, where the follower tracks a virtual target trajectory estimated from the force applied on the leader robot through the operator-designated impedance. This approach was inspired by recent studies in neuroscience that demonstrated the human ability to estimate and use a partner's intended movement during physical interaction \citep{Takagi2017, takagiIndividualsPhysicallyInteracting2019} and the compliant but effective human motor control \citep{gomiEquilibriumPointControlHypothesis1996}. The follower robot is controlled by programming its impedance to be identical to the leader's, and then driving the follower with the leader's virtual target trajectory. IAC enables the user to flexibly set the impedance through their power grasp force, which provides a reasonable estimate of their endpoint stiffness magnitude \citep{takagiEndpointStiffnessMagnitude2020}, adapting it in real-time in accordance with the task at hand. IAC enabled the follower robot to track the leader's movement more accurately than with TIC during both unilateral and bilateral teleoperation.

IAC resembles the state of the art TIC in the sense that both methods enable the operator to modulate the follower's impedance to ensure soft or hard interactions with the remote environment depending on the nature of the task \citep{ajoudaniTeleimpedanceTeleoperationImpedance2012}. However, in TIC the follower's desired trajectory is set to the leader's current position and velocity regardless of follower's impedance. This can cause whiplash, where the follower robot rapidly accelerates and decelerates in a whip-like manner, especially when the operator increases the follower's impedance from a compliant value to a stiff one. This is potentially dangerous as it could  damage objects and people in the remote environment. In contrast, IAC ensures that the follower tracks the leader's target trajectory, which takes into account the impedance controlled by the operator. In doing so, it avoids whiplash when the follower's impedance is adapted rapidly by the operator. Furthermore, the operator's intention is accurately transmitted to the follower robot, ensuring both low and high impedance in response to the operator's desired impedance. {By stabilizing the increase in the follower's impedance for low stiffness and damping values, we ensured stable operation with both TIC and IAC \citep{kronanderStabilityConsiderationsVariable2016}. IAC remained stable even when the follower's stiffness increased during free motion and during physical interactions with the environment}.

The free tracking experiment showed that even when the follower robot's stiffness is greater than 1000\,N/m, IAC outperforms TIC in terms of tracking performance. But where it shines the most is tasks like peg insertion where high compliance can be beneficial to performance. Generally, peg-in-hole is easier for a robot that is compliant in the plane perpendicular to the hole's primary axis. However, with TIC this compliance backfired because it prevented the follower from accurately tracking the leader's movement, especially in its orientation critical to the task. With IAC, high compliance could be achieved whilst also enabling the operator to swiftly correct the follower's position and orientation to rapidly insert the peg into the hole.

IAC vastly expands the range of usable impedance that can be used to control the follower robot with respect to TIC, meaning that the operator has significantly greater control of the follower robot during teleoperation. While a stiffness lower than 600\,N/m was troublesome for TIC with the current setup, with IAC operators could complete tasks quickly even when the follower's stiffness was as low as 40\,N/m. In our experiments, the leader's stiffness was either automatically regulated by the controller for controlled comparisons between TIC and IAC or controlled by the operator using their power grasp force. IAC can also be controlled by electromyograms (EMG) of the arm's muscles, which can be measured and calibrated to obtain a three dimensional estimate of the arm's endpoint impedance ellipse to be used to modify the follower's impedance and the target trajectory. However, EMG signals are very noisy and it takes a significant time to calibrate EMG for each participant, which are factors that need to be considered during practical implementations. In this regard, it may be more practical to estimate the operator's stiffness by measuring the power grasp force as implemented in the current study \citep{takagiEndpointStiffnessMagnitude2020}.

In this study, we used two robots with comparable dynamics to compare TIC and IAC's performance, but it is unclear how large IAC's advantage is in asymmetric setups where the leader and follower robots are of different size and have different dynamics. Pre-compensation of the dynamics difference can be achieved using Eq.\,\re{eq.pre-compensation} with IAC, but its validity should be tested in future work.

\section*{Supplementary Materials}
Video shows a comparison of free tracking and interaction tasks with TIC and IAC, using various impedance ranging from 50\,N/m to 500\,N/m. 

\section*{Appendix}
In the following, we prove that the follower robot will track the leader's movement under the controller in Eq.\,\re{eq.fm}.

Substituting Eq.\,\re{eq.fm} into the robot's model in Eq.\,\re{eq.error}, we obtain the closed-loop dynamics
\bea
\label{eq.closed-loop}
M(x_l)\ddot e+[C(x_l,\dot x_l)+S]\dot e+N e=-\tilde u_l
\eea
as $\hat u_l=u_l+\tilde u_l$ in Eq.\,\re{eq.fm}. From this equation we see that the tracking error $e$ is due to the leader control input estimation error $\tilde u_l$. Therefore, we study the evolution of $e$ by considering $\tilde u_l$, which is determined by the system in Eq.\,\re{eq.state-space-linear} and its observer in Eq.\,\re{eq.observer}, i.e.
\bea
\label{eq.estimation-error}
\dot{\tilde\xi}= \, (A-KH) \, \tilde\xi \, + \, \varepsilon\,, \quad \varepsilon\equiv-\nu -K\mu \, .
\eea
By defining the extended state $\bar\xi \equiv [e,\dot e,\tilde\xi^T]^T$ and combining Eqs.\,\re{eq.closed-loop} and \re{eq.estimation-error}, we obtain the extended system
{\small
\bea
\label{eq.combined-1}
&&\hspace{-7mm}\dot{\bar\xi}= \bar{A} \, \bar\xi + \bar{B} \, \varepsilon,\\
&&\hspace{-7mm}\bar A=\left[
  \begin{array}{ccc}
    0 & 1 & 0 \\
    -M^{-1}N & -M^{-1}(C+S) & M^{-1}\bar H \nb \\
    0 & 0 & A-K H  \\
  \end{array}
\right], \\
&&\hspace{-7mm}\bar B=\left[
  \begin{array}{ccc}
    0 \\
    0 \\
    1 \\
  \end{array}
\right],
~\bar H=\left[
  \begin{array}{cccc}
    0 \\
    0  \\
    0  \\
    1  \\
  \end{array}
\right] \nb
\eea}
The solution of Eq.\,\re{eq.combined-1} is
\bea
\bar\xi = \,e^{\bar A t}\bar\xi(0) \,+ \int_0^{t} \!\!\! e^{\bar A(t-\tau)} \bar B \, \varepsilon(s) \,\, \mbox{d}s
\eea
with the expected value
\bea
\label{eq.E}
E[\bar\xi] = \, e^{\bar A t} E[\bar\xi(0)] \,\, + \int_0^{t} \!\!\! e^{\bar A(t-\tau)} \bar B \, E[\varepsilon(s)] \,\, \mbox{d}s
\eea
Therefore, the stability of the system in Eq.\,\re{eq.combined-1} is determined by the eigenvalues of $\bar{A}$, i.e. the solutions $y$ of the characteristic equation
\bea
[yI-(A-K H)][M y^2+(C+S)y+N]=0 \,.
\eea
The stability of Eq.\,\re{eq.combined-1} is equivalent to the stability of the following two subsystems:
\bea
\label{eq.closed-clean}
&&M\ddot e + (C+S) \, \dot e+ N e=0\\
\label{eq.estimation-error-clean}
&&\dot{\tilde\xi}=(A-KH)\tilde\xi
\eea
The stability of the first subsystem is immediate as Eq.\,\re{eq.leader-closed} is stable. The stability of Eq.\,\re{eq.estimation-error-clean} is ensured by the Kalman filter in Section \ref{sec.intention}. To show that we consider the Lyapunov function candidate
\bea
U= \tilde{\xi}^T \! P_v \, \tilde{\xi}\,, \quad P_v \equiv P^{-1} \,.
\eea
From the Riccati equation in Eq.\,\re{eq.riccati} we get
\bea
\label{eq.riccati-inverse}
P_v A+ A^T P_v-H^TR^{-1} H+P_v Q P_v=-\dot P_v \, .
\eea
The time derivative of $U$ is
\bea
\dot U = \, \tilde \xi^T \dot P_v\tilde\xi \, + 2\,\tilde \xi^T P_v\dot{\tilde\xi}\,.
\eea
According to the definition of $K$ in Eqs.\,\re{eq.K} and \re{eq.estimation-error-clean}, we have
\bea
\dot{\tilde\xi}=\,(A-P H^TR^{-1}H)\, \tilde{\xi} \,.
\eea
It follows that
\bea
\label{eq.dot-tilde-varphi}
2\,\tilde \xi^T P_v\dot{\tilde\xi} \, = \, \tilde \xi^T(P_v A+A^T P_v-2H^TR^{-1}H) \tilde{\xi} \, .
\eea
With Eqs.\,\re{eq.riccati-inverse} and \re{eq.dot-tilde-varphi}, we then obtain
\bea
\label{eq.V2dot}
\dot U = - (P_v\tilde{\xi})^T Q \, P_v\tilde{\xi} -
(H \tilde{\xi})^T R^{-1} H \tilde{\xi} \, \leq \, 0
\eea
which indicates that the system in Eq.\,\re{eq.estimation-error-clean} is stable. From the definition of $H$ in Eq.\,\re{eq.measurement}, we know that  the estimation error of observable states $[\tilde x, \dot{\tilde x},\tilde u_h]$ is asymptotically stable, except the unobservable state $\tilde\theta$. Therefore, the first term in Eq.\,\re{eq.E} vanishes for $t\rightarrow \infty$, except $E[\tilde\theta(0)]$ which is bounded. In the second term, $E[\varepsilon(s)]=0$ because $E[\nu]\,=\,E[\mu]\,=\,0$. According to the definition of the state $\bar\xi=[e,\dot e,\tilde\xi^T]^T$, we have $E[\dot e]\rightarrow 0$, $E[e]\rightarrow 0$.

\ifCLASSOPTIONcaptionsoff
  \newpage
\fi

\bibliographystyle{plainnat}
\bibliography{MyLibrary,YananLibrary} 

@article{ajoudaniTeleimpedanceTeleoperationImpedance2012,
  title = {Tele-Impedance: {{Teleoperation}} with Impedance Regulation Using a Body--Machine Interface},
  shorttitle = {Tele-Impedance},
  author = {Ajoudani, Arash and Tsagarakis, Nikos and Bicchi, Antonio},
  year = 2012,
  month = nov,
  journal = {The International Journal of Robotics Research},
  volume = {31},
  number = {13},
  pages = {1642--1656},
  publisher = {SAGE Publications Ltd STM},
  issn = {0278-3649},
  urldate = {2021-02-15},
  langid = {english}
}

@inproceedings{andersonBilateralControlTeleoperators1988,
  title = {Bilateral Control of Teleoperators with Time Delay},
  booktitle = {Proceedings of the 1988 {{IEEE International Conference}} on {{Systems}}, {{Man}}, and {{Cybernetics}}},
  author = {Anderson, R.J. and Spong, M.W.},
  year = 1988,
  month = aug,
  volume = {1},
  pages = {131--138},
  urldate = {2024-02-05}
}

@article{edenHapticCommunicationCentral2024,
  title = {During Haptic Communication, the Central Nervous System Compensates Distinctly for Delay and Noise},
  author = {Eden, Jonathan and Ivanova, Ekaterina and Burdet, Etienne},
  year = 2024,
  month = nov,
  journal = {PLOS Computational Biology},
  volume = {20},
  number = {11},
  pages = {e1012037},
  publisher = {Public Library of Science},
  issn = {1553-7358},
  urldate = {2025-12-08},
  langid = {english}
}

@article{fundaTeleprogrammingDelayinvariantRemote1992,
  title = {Teleprogramming: {{Toward}} Delay-Invariant Remote Manipulation},
  shorttitle = {Teleprogramming},
  author = {Funda, Janez and Lindsay, Thomas S. and Paul, Richard P.},
  year = 1992,
  month = jan,
  journal = {Presence: Teleoperators and Virtual Environments},
  volume = {1},
  number = {1},
  pages = {29--44},
  issn = {1054-7460}
}

@article{gomiEquilibriumpointControlHypothesis1996,
  title = {Equilibrium-Point Control Hypothesis Examined by Measured Arm Stiffness during Multijoint Movement},
  author = {Gomi, Hiroaki and Kawato, Mitsuo},
  year = 1996,
  month = apr,
  journal = {Science},
  volume = {272},
  number = {5258},
  pages = {117--120},
  issn = {0036-8075, 1095-9203},
  urldate = {2018-09-11},
  copyright = {\copyright{} 1996 American Association for the Advancement of Science},
  langid = {english},
  pmid = {8600521}
}

@article{kronanderStabilityConsiderationsVariable2016,
  title = {Stability {{Considerations}} for {{Variable Impedance Control}}},
  author = {Kronander, Klas and Billard, Aude},
  year = 2016,
  month = oct,
  journal = {IEEE Transactions on Robotics},
  volume = {32},
  number = {5},
  pages = {1298--1305},
  issn = {1941-0468},
  urldate = {2024-07-09}
}

@article{lawrenceStabilityTransparencyBilateral1993,
  title = {Stability and Transparency in Bilateral Teleoperation},
  author = {Lawrence, D.A.},
  year = 1993,
  month = oct,
  journal = {IEEE Transactions on Robotics and Automation},
  volume = {9},
  number = {5},
  pages = {624--637},
  issn = {2374-958X},
  urldate = {2024-02-05}
}

@article{miallAdaptationVisualFeedback2006,
  title = {Adaptation to Visual Feedback Delays in Manual Tracking: Evidence against the {{Smith Predictor}} Model of Human Visually Guided Action},
  shorttitle = {Adaptation to Visual Feedback Delays in Manual Tracking},
  author = {Miall, R. C. and Jackson, J. K.},
  year = 2006,
  month = jun,
  journal = {Experimental Brain Research},
  volume = {172},
  number = {1},
  pages = {77--84},
  issn = {1432-1106},
  urldate = {2020-02-13},
  langid = {english}
}

@article{nankakuMaximumAcceptableCommunication2022,
  title = {Maximum Acceptable Communication Delay for the Realization of Telesurgery},
  author = {Nankaku, Akitoshi and Tokunaga, Masanori and Yonezawa, Hiroki and Kanno, Takahiro and Kawashima, Kenji and Hakamada, Kenichi and Hirano, Satoshi and Oki, Eiji and Mori, Masaki and Kinugasa, Yusuke},
  year = 2022,
  journal = {PloS One},
  volume = {17},
  number = {10},
  pages = {e0274328},
  issn = {1932-6203},
  langid = {english},
  pmcid = {PMC9536636},
  pmid = {36201429}
}

@article{niemeyerStableAdaptiveTeleoperation1991,
  title = {Stable Adaptive Teleoperation},
  author = {Niemeyer, G. and Slotine, J.-J.E.},
  year = 1991,
  month = jan,
  journal = {IEEE Journal of Oceanic Engineering},
  volume = {16},
  number = {1},
  pages = {152--162},
  issn = {1558-1691},
  urldate = {2025-12-08}
}

@article{panNewPredictiveApproach2006,
  title = {A New Predictive Approach for Bilateral Teleoperation with Applications to Drive-by-Wire Systems},
  author = {Pan, Ya-Jun and {Canudas-de-Wit}, C. and Sename, O.},
  year = 2006,
  month = dec,
  journal = {IEEE Transactions on Robotics},
  volume = {22},
  number = {6},
  pages = {1146--1162},
  issn = {1552-3098},
  urldate = {2024-02-05}
}

@article{panzirschExploringPlanetGeology2022,
  title = {Exploring Planet Geology through Force-Feedback Telemanipulation from Orbit},
  author = {Panzirsch, Michael and Pereira, Aaron and Singh, Harsimran and Weber, Bernhard and Ferreira, Edmundo and Gherghescu, Andrei and Hann, Lukas and {den Exter}, Emiel and {van der Hulst}, Frank and Gerdes, Levin and Cencetti, Leonardo and Wormnes, Kjetil and Grenouilleau, Jessica and Carey, William and Balachandran, Ribin and Hulin, Thomas and Ott, Christian and Leidner, Daniel and {Albu-Sch{\"a}ffer}, Alin and Lii, Neal Y. and Kr{\"u}ger, Thomas},
  year = 2022,
  month = apr,
  journal = {Science Robotics},
  volume = {7},
  number = {65},
  pages = {eabl6307},
  publisher = {American Association for the Advancement of Science},
  urldate = {2024-02-05}
}

@article{takagiAnalogousAdaptationsSpeed2020,
  title = {Analogous Adaptations in Speed, Impulse and Endpoint Stiffness When Learning a Real and Virtual Insertion Task with Haptic Feedback},
  author = {Takagi, Atsushi and De Magistris, Giovanni and Xiong, Geyun and Micaelli, Alain and Kambara, Hiroyuki and Koike, Yasuharu and Savin, Jonathan and Marsot, Jacques and Burdet, Etienne},
  year = 2020,
  month = dec,
  journal = {Scientific Reports},
  volume = {10},
  number = {1},
  pages = {22342},
  publisher = {Nature Publishing Group},
  issn = {2045-2322},
  urldate = {2020-12-20},
  copyright = {2020 The Author(s)},
  langid = {english}
}

@article{takagiEndpointStiffnessMagnitude2020,
  title = {Endpoint Stiffness Magnitude Increases Linearly with a Stronger Power Grasp},
  author = {Takagi, A. and Xiong, G. and Kambara, H. and Koike, Y.},
  year = 2020,
  month = jan,
  journal = {Scientific Reports},
  volume = {10},
  number = {1},
  pages = {1--9},
  issn = {2045-2322},
  urldate = {2020-01-16},
  copyright = {2020 The Author(s)},
  langid = {english}
}

@article{takagiIndividualsPhysicallyInteracting2019,
  title = {Individuals Physically Interacting in a Group Rapidly Coordinate Their Movement by Estimating the Collective Goal},
  author = {Takagi, Atsushi and Hirashima, Masaya and Nozaki, Daichi and Burdet, Etienne},
  editor = {Diedrichsen, J{\"o}rn and Ivry, Richard B and Diedrichsen, J{\"o}rn and Giese, Martin A},
  year = 2019,
  month = feb,
  journal = {eLife},
  volume = {8},
  pages = {e41328},
  issn = {2050-084X},
  urldate = {2019-04-09}
}

@article{yokokohjiBilateralControlMasterslave1994,
  title = {Bilateral Control of Master-Slave Manipulators for Ideal Kinesthetic Coupling--Formulation and Experiment},
  author = {Yokokohji, Y. and Yoshikawa, T.},
  year = 1994,
  month = oct,
  journal = {IEEE transactions on robotics and automation: a publication of the IEEE Robotics and Automation Society},
  volume = {10},
  number = {5},
  pages = {605--620},
  issn = {1042-296X},
  langid = {english},
  pmid = {11539289}
}

@article{OUYANG2006,
title = "An adaptive switching learning control method for trajectory tracking of robot manipulators",
journal = "Mechatronics",
volume = "16",
number = "1",
pages = "51 - 61",
year = "2006",
issn = "0957-4158",
author = "P.R. Ouyang and W.J. Zhang and Madan M. Gupta"
}

@ARTICLE{HircheS12PI,
 author={S. Hirche and M. Buss},
 title={Human-Oriented Control for Haptic Teleoperation},
 Journal={Proceedings of the IEEE},
 volume={100},
 issue={3},
 year={2012},
 pages="623-647",
 }

@article{PassenbergC10M,
title = "A survey of environment-, operator-, and task-adapted controllers for teleoperation systems",
journal = "Mechatronics ",
volume = "20",
number = "7",
pages = "787 - 801",
year = "2010",
author = "C. Passenberg and A. Peer and M. Buss",
}

@ARTICLE{Tobergte09,
 author={A. Tobergte and R. Konietschke and G. Hirzinger},
 title={{Planning and control of a teleoperation system for research in minimally invasive robotic surgery}},
 Journal={{IEEE International Conference on Robotics and Automation (ICRA)}},
 year={2009},
 pages="4225-4232",
 }

@article{Takagi2017,
  title={Physically interacting individuals estimate the partner’s goal to enhance their movements},
  author={Atsushi Takagi and Gowrishankar Ganesh and Toshinori Yoshioka and Mitsuo Kawato and Etienne Burdet},
  journal={Nature Human Behaviour},
  year={2017},
  volume={1}
}

@article{Ganesh2014,
  title={Two is better than one: Physical interactions improve motor performance in humans},
  author={G. Ganesh and A. Takagi and R. Osu and T. Yoshioka and M. Kawato and E. Burdet},
  journal={Scientific Reports},
  year={2014},
  volume={4},
  number={3824},
}

@INPROCEEDINGS{Xia2011,
author={T. Xia and A. Kapoor and P. Kazanzides and R. Taylor},
booktitle={IEEE/RSJ International Conference on Intelligent Robots and Systems (IROS)},
title={A constrained optimization approach to virtual fixtures for multi-robot collaborative teleoperation},
year={2011},
volume={},
number={},
pages={639-644},
ISSN={2153-0866},}

@article{Li2019,
author = {Y. Li and G. Carboni and F. Gonzalez and D. Campolo and E. Burdet},
title ={Differential game theory for versatile physical human–robot interaction},
journal = {Nature Machine Intelligence},
volume = {1},
number = {},
pages = {36-43},
year = {2019},
}

@article{Takagi2018a,
author = {Takagi, Atsushi and Usai, Francesco and Ganesh, Gowrishankar and Sanguineti, Vittorio and Burdet, Etienne},
issn = {1553-7358},
journal = {PLOS Computational Biology},
month = {mar},
number = {3},
pages = {e1005971},
publisher = {Public Library of Science},
title = {{Haptic communication between humans is tuned by the hard or soft mechanics of interaction}},
volume = {14},
year = {2018}
}

@article{Li2014,
author = {Li, Yanan and Ge, Shuzhi Sam},
issn = {1083-4435},
journal = {IEEE/ASME Transactions on Mechatronics},
number = {3},
pages = {1007--1014},
title = {{Human--robot collaboration based on motion intention estimation}},
volume = {19},
year = {2014}
}

@INPROCEEDINGS{Atashzar2012,

  author={S. F. {Atashzar} and I. G. {Polushin} and R. V. {Patel}},

  booktitle={IEEE/RSJ International Conference on Intelligent Robots and Systems (IROS)}, 

  title={Networked teleoperation with non-passive environment: Application to tele-rehabilitation}, 

  year={2012},

  volume={},

  number={},

  pages={5125-5130},}

@ARTICLE{Yoon2004,

  author={ {Woo-Keun Yoon} and T. {Goshozono} and H. {Kawabe} and M. {Kinami} and Y. {Tsumaki} and M. {Uchiyama} and M. {Oda} and T. {Doi}},

  journal={IEEE Transactions on Robotics and Automation}, 

  title={Model-based space robot teleoperation of ETS-VII manipulator}, 

  year={2004},

  volume={20},

  number={3},

  pages={602-612},}

@article{Frizera2010,
author = {Frizera, Anselmo and Gallego, Juan and Rocon, Eduardo and Pons, José and Ceres, R.},
year = {2010},
month = {08},
pages = {37},
title = {Extraction of user's navigation commands from upper body force interaction in walker assisted gait},
volume = {9},
journal = {Biomedical Engineering Online},
doi = {10.1186/1475-925X-9-37}
}

@article{Vanderborght2013,
title = "Variable impedance actuators: A review",
journal = "Robotics and Autonomous Systems",
volume = "61",
number = "12",
pages = "1601 - 1614",
year = "2013",
issn = "0921-8890",
doi = "https://doi.org/10.1016/j.robot.2013.06.009",
url = "http://www.sciencedirect.com/science/article/pii/S0921889013001188",
author = "B. Vanderborght and A. Albu-Schaeffer and A. Bicchi and E. Burdet and D.G. Caldwell and R. Carloni and M. Catalano and O. Eiberger and W. Friedl and G. Ganesh and M. Garabini and M. Grebenstein and G. Grioli and S. Haddadin and H. Hoppner and A. Jafari and M. Laffranchi and D. Lefeber and F. Petit and S. Stramigioli and N. Tsagarakis and M. {Van Damme} and R. {Van Ham} and L.C. Visser and S. Wolf"
}

@article{Takagi-Li2020,
author = {Atsushi Takagi and Yanan Li and Etienne Burdet},
title ={Flexible Assimilation of Human’s Target for
Versatile Human-robot Physical Interaction},
journal = {IEEE Transactions on Haptics},
volume = {14},
number = {2},
pages = {421-431},
year = {2020},
}

@article{Laghi2020,
author = {Marco Laghi and Arash Ajoudani and Manuel G. Catalano and Antonio Bicchi},
title ={Unifying bilateral teleoperation and tele-impedance for enhanced user experience},
journal = {The International Journal of Robotics Research},
volume = {39},
number = {4},
pages = {514-539},
year = {2020},
doi = {10.1177/0278364919891773}
}

\end{document}